\pdfoutput=1

\documentclass[11pt]{article}

\usepackage[final]{coling}

\usepackage{times}
\usepackage{latexsym}

\usepackage[T1]{fontenc}

\usepackage[utf8]{inputenc}

\usepackage{microtype}

\usepackage{inconsolata}

\usepackage{graphicx}

\usepackage{amsmath}

\usepackage{enumitem}
\usepackage{multirow}
\usepackage{booktabs}
\usepackage{algorithm}
\usepackage{algorithmic}
\usepackage{xcolor}
\usepackage{subfigure}
\usepackage{CJKutf8}

\title{\textit{Teaching According to Talents!}\\Instruction Tuning LLMs with Competence-Aware Curriculum Learning}

\author{
 \textbf{Yangning Li\textsuperscript{1,2,}\thanks{$\,$Equal Contribution.}},
 \textbf{Tingwei Lu\textsuperscript{1,}$^{*}$},
 \textbf{Yinghui Li\textsuperscript{1}},
 \textbf{Yankai Chen\textsuperscript{3}},
 \textbf{Wei-Chieh Huang\textsuperscript{4}},
 \\
 \textbf{Wenhao Jiang\textsuperscript{5}},
 \textbf{Hui Wang\textsuperscript{2}},
 \textbf{Hai-Tao Zheng\textsuperscript{1,2}\thanks{Corresponding author: zheng.haitao@sz.tsinghua.edu.cn}},
 \textbf{Philip S.Yu\textsuperscript{4}},
\\
 \textsuperscript{1}Tsinghua University, 
 \textsuperscript{2}Peng Cheng Laboratory, 
 \textsuperscript{3}Cornell University \\
 \textsuperscript{4}University of Illinois Chicago,
 \textsuperscript{5}Guangming Laboratory
}

\begin{document}
\maketitle
\begin{abstract}
Efficient instruction tuning aims to enhance the ultimate performance of large language models (LLMs) trained on a given instruction dataset. Curriculum learning as a typical data organization strategy has shown preliminary effectiveness in instruction tuning. However, current curriculum tuning methods suffer from the curriculum rigidity, since they rely solely on static heuristic difficulty metrics. These methods fail to adapt to the evolving capabilities of models during training, resulting in a fixed and potentially sub-optimal learning trajectory. To address the issue, \textbf{C}ompetence-\textbf{A}ware \textbf{M}ulti-\textbf{P}erspective c\textbf{U}rriculum in\textbf{S}truction tuning framework termed \textbf{CAMPUS} is proposed. CAMPUS offers several advantages: (1) Dynamic selection for sub-curriculum. (2) Competency-aware adjustment to the curriculum schedule. (3) Multiple difficulty-based scheduling. Extensive experiments\footnote{The code will be open source.} prove the superior performance of CAMPUS, compared to other state-of-the-art baselines for efficient instruction tuning.
\end{abstract}

\section{Introduction}


Instruction tuning \cite{zhang2023instruction,zhao2023survey, yu2024seqgpt} aligns large language models (LLMs) \cite{touvron2023llama,bai2023qwen, li2024llms, huang2023lateval, li2024benchmarking, li2024rethinking, xu2024let} with human preferences, enhancing their effectiveness across real-world tasks. Some studies \cite{zhou2024lima,wang2023self} have highlighted the importance of instruction data management, including data quality curation and training strategy. This has fueled interest in \textit{efficient instruction tuning} \cite{wang2024survey,wang2023data}, which aims \textbf{to maximize LLM performance trained on a given instruction dataset}. Among various strategies, optimizing training order has emerged as a key research focus.

Curriculum learning \cite{bengio2009curriculum}, as a typical data order organization strategy that mimics the learning process of human education, has shown preliminary effectiveness in instruction tuning. The essential idea is to train LLMs with instruction data in a progression from easy to difficult, thereby accelerating model convergence and achieving a higher performance upper bound. The key to curriculum learning lies in the design of difficulty metrics for instruction data. For instance, Tree-Instruct \cite{zhao2024tree} treats instruction data as a semantic tree and measures difficulty based on the number of tree nodes. Referring to the expert-designed educational framework, \citet{lee2024instruction} utilizes ChatGPT to synthesize the instruction dataset CORGI with increasing difficulty, and then train the LLMs sequentially.

\begin{figure}[t]
\centering
\scalebox{1}{
\includegraphics[width=0.48\textwidth]{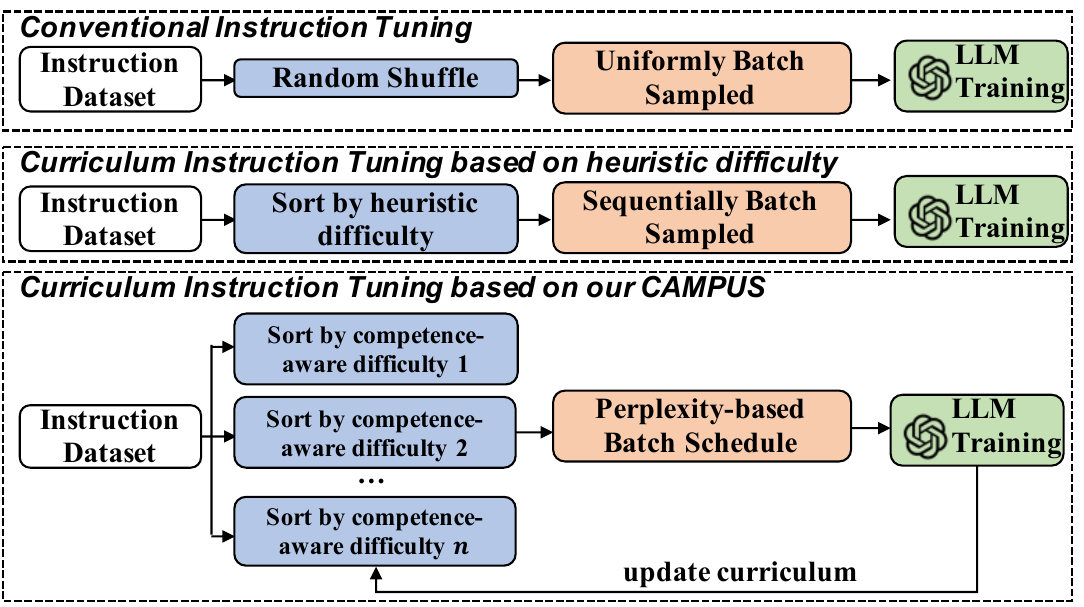}}
\caption{Comparison of different curriculum instruction tuning.}
\label{fig:intro}
\end{figure}

Despite these advancements, \textbf{previous difficulty metrics are often static and predefined with human intuition}. These metrics fail to gauge the difficulty variations of the same data for different LLMs or at different learning stages. Consequently, as shown in Figure \ref{fig:intro}, curriculum instruction tuning based on them lacks the flexibility to adjust curriculum for different LLMs or as LLMs evolve. The rigid curriculum schedule result in a sub-optimization in instruction tuning. As educator John Dewey said, ``if we teach today's students as we taught yesterday's, we rob them of tomorrow," underscoring the necessity of adaptable learning strategies in human cognition. However, it remains unclear for us how to customize a ``suitable'' instruction curriculum for a specific LLM.

In light of the above issues, we seek to develop a data order organization strategy that customizes a dynamic curriculum for different LLMs or LLM at different training phases, based on which we can improve the training efficiency of instruction dataset. To this end, a \textbf{C}ompetence-\textbf{A}ware \textbf{M}ulti-\textbf{P}erspective c\textbf{U}rriculum in\textbf{S}truction tuning framework termed \textbf{CAMPUS} is proposed. Initially, CAMPUS independently sorts instruction data using multiple difficulty metrics, each representing a curriculum schedule developed from distinct perspectives. At each training step, CAMPUS assesses the LLM competence and inherent data difficulty, selecting the sub-curriculum from the various schedules that poses the minimum perplexity, i.e., sub-curriculum that can be relatively easily comprehended and internalized by current LLM.

CAMPUS is compatible with arbitrary quantities of difficulties. Besides the conventional heuristic difficulty, we introduce competence-aware difficulty based on a lightweight scoring model, which is trained with adversarial learning. It considers both LLM parameters and instruction data as joint inputs to assess the instruction data's difficulty for the current LLM. As the LLM's training progresses, the competence-aware difficulty consistently evolves, allowing the curriculum based on the reward model to dynamically adjust.


Compared to existing curriculum instruction tuning methods, CAMPUS offers the following advantages:
(1) \textbf{Dynamic Selection for Sub-Curriculum}: It continuously evaluates the learning state and progress of the model, selecting the most appropriate sub-curriculum to match the current learning needs.
(2) \textbf{Competency-Aware Adjustment to the Curriculum Schedule}: CAMPUS adjusts the curriculum schedule based on real-time assessments of the model’s competencies, ensuring that the curriculum aligns with the model's evolving strengths and weaknesses.
(3) \textbf{Multiple Difficulty-Based Scheduling}: Unlike methods that rely on a single static metric, CAMPUS utilizes multiple difficulty metric, holding a more comprehensive perspective of the data difficulty.

LLMs from different families are equipped with CAMPUS to demonstrate their effectiveness, including BLOOMZ and LLaMA. In experiments, we utilized a mixed instruction dataset derived from Code Alpaca, GSM8K, and ShareGPT as the original training dataset $D$. Various types of efficient instruction methods, including data selection and training order optimization, are served as baseline models to enhance training efficiency with $D$. We conduct an extensive evaluation on three benchmarks: GSM8K for mathematical reasoning, HumanEval for coding, and MT-Bench for general language understanding. CAMPUS consistently outperforms state-of-the-art methods by an average of 7.0\%. Additionally, as a pluggable training strategy, CAMPUS can be integrated with other data selection methods to further enhance the efficiency of instruction tuning.

\section{Related Work}
\subsection{Efficient Instruction Tuning}
Instruction Tuning \cite{zhang2023instruction,zhao2023survey, li2025mdit, chen2025dast, wang2024exploring} aims to align the pre-trained LLM abilities towards human preference, which enables LLMs to quickly adapt to specific domains or acquire specialized skills \cite{kuang2025natural, qin2024multilingual,li2023effectiveness, li2022past, qin2024large, du2024llms,li2025one,li2025refine, kuang2025express, qingsong2025raise}. 
Some recent research \cite{zhou2024lima,wang2023self} reveals that only a fairly small amount of high-quality data is needed to align large models well with human preferences, in contrast to traditional task-specific fine-tuning \cite{kenton2019bert, ma2022linguistic, li2022learning, li2022contrastive} or pre-training \cite{NEURIPS2020_1457c0d6,dong2022survey, liu2022we} where data quantity is crucial. This finding has motivated the AI research community to dedicate on \textit{efficient instruction tuning} \cite{wang2024survey,wang2023data}, including data and training efficiency. 
The common objective of various methods is to \textbf{enhance the ultimate performance of LLMs trained on a given instruction dataset $D$}.

The essence of the data-efficient instruction tuning \cite{li2024quantity,dong-etal-2024-abilities,liu2024what} lies in selecting a high-quality subset $D_{sub}\subseteq D$ from original datasets. Previous work filtered data from three dimensions: quality, diversity, and complexity. For quality assessment \cite{chenalpagasus}, powerful LLMs such as ChatGPT or fine-tuned LLMs are frequently employed as quality evaluators. Diversity \cite{wu2023self} is often measured using heuristic metrics like ROUGE-L similarity. Additionally, some works also endeavor to quantify and evaluate instruction complexity. For instance, \#InsTag \cite{lu2023instag} proposes to quantify instruction complexity using the number of fine-grained tags generated by ChatGPT. DEITA \cite{liu2024what} introduces a comprehensive method that simultaneously considers all three aspects, enabling automatic data selection.


\subsection{Curriculum Instruction Tuning}
A representative way for training-efficient instruction tuning is to organize the order of training data. Curriculum learning \cite{bengio2009curriculum, zhang2023contextual, li2025correct, yuan2023curriculum, xing2024mitigating, zhang2025loss, li2024mesed, li2023vision}, as a data organization strategy that emulates the human brain's learning process, has demonstrated effectiveness in instruction tuning. Its core principle involves arranging learning data from simple to hard, and the key is the design of difficulty metric. \citet{zhao2024tree} and \citet{sun2024conifer} have employed the number of nodes in semantic trees and evaluations by ChatGPT, respectively, to gauge difficulty and create learning schedules. \citet{lee2024instruction} leverages expert-designed educational frameworks as basic curriculum outline and then synthesizes easy-to-hard instruction dataset CORGI. However, these rigid heuristic difficulties hindered dynamic adjustment of curriculum schedules.

\section{Method}
\subsection{Overall Framework}
\begin{figure}[t]
\centering
\scalebox{0.9}{
\includegraphics[width=0.45\textwidth]{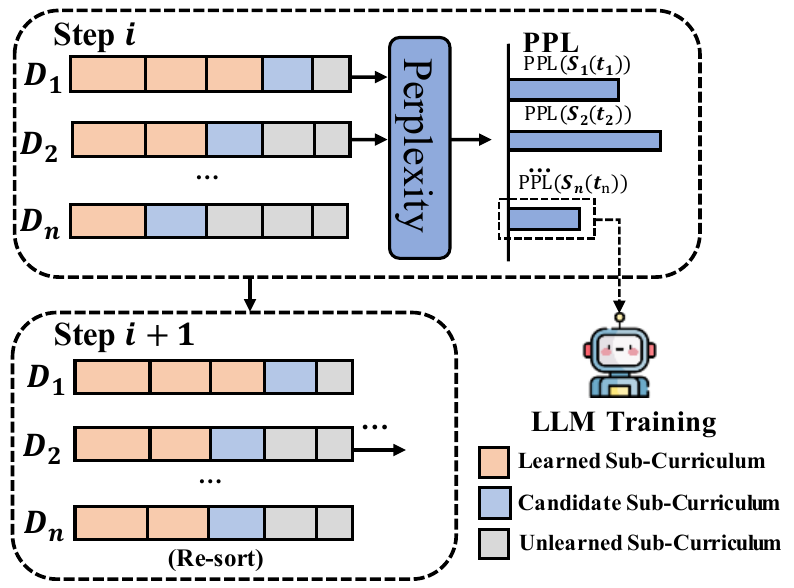}}
\caption{Framework of our CAMPUS to demonstrate how to schedule the sub-curricula.}
\label{fig:method}
\vspace{-0.5cm}
\end{figure}

Previous curriculum instruction tuning typically relied on static difficulty metrics, which limited their ability to assess the difficulty variations of the same data for different LLMs, and also hindered dynamic adjustment of learning schedules based on LLMs' mastery of the training data during the learning processes. To address these issues, we propose a \textbf{C}ompetence-\textbf{A}ware \textbf{M}ulti-\textbf{P}erspective c\textbf{U}rriculum in\textbf{S}truction tuning framework, termed \textbf{CAMPUS}. The overall framework is illustrated in Figure \ref{fig:method}. CAMPUS first sorts the instruction dataset individually based on $n$ difficulty metrics, and then dynamically programs the training batch during the training process utilizing a perplexity-based curriculum scheduler. As for the difficulty metrics, in addition to the heuristic metrics as in past work, we introduce novel model competence-aware metrics. Besides the inherent data difficulty, these metrics also account for LLMs' proficiency in data and their learning capacity, enabling flexible adjustments to the curriculum schedule.
The details of curriculum scheduler and metrics will be provided in Section \ref{sec:multi_schedule} and \ref{sec:metrics}.

\begin{algorithm}[t]
\small
	\renewcommand{\algorithmicrequire}{\textbf{Input:}}
	\renewcommand{\algorithmicensure}{\textbf{Output:}}
	\caption{\small{Overall process of \textbf{CAMPUS}.} }
	\begin{algorithmic}[1]
		\REQUIRE Training instruction dataset $D$, $i \in \{1,2,3,4\}$.
		\ENSURE A LLM with \textbf{CAMPUS} learning.
		\STATE Compute four difficulties $d_i$ for each data sample in $D$; 
		\STATE Initial Sort $D$ based each difficulty of every sample, resulting in $D_i$ (i.e., $D_1$, $D_2$, $D_3$, $D_4$);
		\FOR{$i = 1, 2, 3, 4$}
		\STATE $t_i = 1$; Initialize the learning scope from $i^{th}$ perspective, $s_i(1)$, by Eq.~(\ref{eq:scope});
		\STATE {Compute the perplexity (PPL) on top $s_i(1)$ portion, PPL($S_i(1)$), by Eq.~(\ref{eq:ppl})};
		\ENDFOR
		\REPEAT
		\STATE {$j = \underset{i}{\arg\max} (\text{PPL}(S_i(t_i)))$};
		\STATE Train the LLM with the $S_j(t_j)$;
            \IF{$j^{th}$ difficulty is competence-aware}
  		\STATE  Re-sort the curriculum order after $s_j(t_j)$;
            \ENDIF
		\STATE $t_j \leftarrow t_j + 1$;
		\STATE Update the learning scope from $j^{th}$ perspective, $S_j(t_j)$, by Eq.~(\ref{eq:scope}); Update
  the candidate training batch from the scope $S_j(t_j)$ portions of $D_j$;
		\STATE { Compute the PPL of LLM on $S_j(t_j)$ portion , PPL($S_j(t_j)$), by Eq.~(\ref{eq:ppl})};
		\UNTIL LLM converges.
	\end{algorithmic}
 \label{alg:1}
\end{algorithm}

\subsection{Dynamic Curriculum Scheduler}
\label{sec:multi_schedule}
For the given $n$ difficulties, CAMPUS separately organizes the original instruction dataset into $n$ easy-to-hard curriculum schedules $\{D_1,D_2,\ldots,D_n\}$, which can be considered as designed from distinct pedagogical perspectives. The primary objective is to intricately devise a sub-curriculum program strategy across these schedules to enhance the training of LLM, tailored to the LLM capabilities and sub-curriculum complexity. A straightforward method might involve selecting sub-curricula in a sequential format (i.e., $1 \rightarrow 2 \rightarrow...\rightarrow n \rightarrow 1$), with the expectation that the overall curriculum order would still adhere to a progressive difficulty gradient. However, this method proves ineffective as it does not account for the incomparability of difficulty across heterogeneous measurement scales of different schedules. Hence, a dynamic curriculum scheduler that leverages perplexity metrics is proposed. First, we define the learning scope $s(t)$ for a particular curriculum schedule at training step $t$ as follows:
\begin{equation}
\label{eq:scope}
\small
    s(t)=\min(1, \sqrt[p]{t \frac{1-s(1)^p}{T}+s(1)^p}), t=2,3,... ,T
\end{equation}
where $s(1)=0.01$ represents the initial learning scope, $p=2$ controls the curriculum progression rate, and $T$ represents the total training steps. Each curriculum $D_i$ is segmented into $T$ sub-curricula. The sub-curriculum range $S_i(t_j)$ at step $t_j$ is denoted as by $[s(t_{j-1}), s(t_j)]$, which is utilized as a training batch. Given that $s(t)$ is an exponential function proportional to $\sqrt[p]{\frac{t}{T}}$, its growth decelerates with increasing $t$. In other words, the learning granularity becomes finer as difficulty of curriculum escalates, akin to human educational practices. For each candidate training batch (sub-curriculum) derived from these schedules, the scheduler calculates the perplexity (PPL), defined as:
\begin{equation}
\label{eq:ppl}
\small
    \operatorname{PPL}(S_i(t_i))=\frac{\sum_{X\in S_i(t_i)}\sqrt[N]{\prod_{m=1}^N \frac{1}{P(w_m|w_1,w_2,... ,w_{m-1})}}}{|S_i(t_i)|}
\end{equation}
where $X=\{w_1,w_2,... ,w_N\}$ represents a data sample in $S_i(t_i)$. The PPL measures the model's confusion with real data samples. A training sample with a higher PPL indicates that the LLM has not mastered the basic ability to comprehend it and needs deferred learning. Consequently, the scheduler dynamically selects the training batch with the minimum PPL among the candidates, maintaining an orderly progression of overall difficulty. After that, the curriculum order is updated if the selected metric is competence-aware, and the process iteratively proceeds to the next sub-curriculum selection. Algorithm \ref{alg:1} presents the entire process.

\subsection{Competence-Aware Difficulty Metrics}
\label{sec:metrics}
In our experiments, we employed four difficulty metrics (i.e., $n$=4), of which loss and reward score are competence-aware. These metrics not only assess the inherent difficulty of the instruction data but also adjust according to the LLM's proficiency with the skills associated with the instruction data. In other words, they serve as difficulty indicator functions $d_i=f_i\left(\mathrm{inst},\theta_{\mathrm{LLM}}\right)$ with data and model parameter as inputs. Additionally, we introduced heuristic difficulty metrics data length and textual lexical diversity, considering their empirical effectiveness. \textbf{Each metric is viewed as capturing the curriculum from a different perspective, thereby reducing the potential bias that could arise from relying on a single metric.}
More meaningful metrics can be integrated into CAMPUS in the future.

\noindent\textbf{Data Length} $d_1$ The data length is well-suited as a heuristic difficulty metric, aligning with intuitive human assessments: shorter instruction data are typically easier for LLMs to comprehend and learn. Consequently, we employed the total length of the instruction and output tokens, denoted as $\operatorname{len}(x+y)$, as $d_1$.

\noindent\textbf{Measure of textual lexical diversity (MTLD)} $d_2$ Lexical diversity refers to the variety of different words employed in a text, which serves as a crucial factor in studies of linguistic complexity. MTLD \cite{mccarthy2010mtld} is a widely used quantitative measure for assessing lexical diversity. It is calculated as the mean length of sequential token lists in a text that maintains a given type-token ratio threshold. We selected MTLD as a metric due to its length-insensitivity, since text length does not always accurately reflect the data complexity, e.g., redundancies. MTLD can be considered as a complementary perspective on instruction complexity that is distinct from data length.

\noindent\textbf{Data Loss} $d_3$ The data difficulty as perceived by LLMs depends not only on the data's inherent complexity, but also on the LLMs' comprehension abilities. Therefore, in addition to the two heuristics mentioned above, we introduce a simple, model competence-aware difficulty metric, data loss, as $d_3$. Data loss indirectly quantifies problem-solving capacity of LLM by measuring the difference between the predicted content and the ground-truth value, which is defined as:
$$
d_3=L(x,y,\theta_{LLM})=-\sum_{t=1}^N log p_{\theta_{LLM}}(y_t|x,y_{<t})
$$
where $N$ is the length of the output text sequence and $\theta_{LLM}$ is the LLM parameter. A higher loss suggests that LLM struggles with tasks related to specific instruction data.

\noindent\textbf{Competence-aware Data Score} $d_4$ When assessing the data difficulty under specific model, data loss typically focuses solely on the difference between the model's output and the actual results, often neglecting the inherent data complexity and inner state of model. Consequently, we propose a model competent-aware scoring model $R$ that takes both the data and model state as joint input, and outputs the data difficulty, denoted as $d_4$. To train the model $R$, we assign labels to each piece of instruction data. Empirically, models typically perceive data they have been trained on as straightforward, labeling these as 0; conversely, data they have not encountered are considered challenging, thus labeled as 1. 

We commence by randomly shuffling the dataset $D$ (We used the same training data in Appendix \ref{sec:traindata_appendix} to train the scoring model) and equally dividing it into $n$ portions,  $D_1, D_2, \ldots, D_n$. At the $i^{th}$ epoch of training $R$, we choose the dataset pair $(D_i, D_{i+1})$ to process. We update the LLM parameters from the last epoch using dataset $D_i$:
\begin{equation}
\small
\theta_{LLM}^{(i)}=SFT_{x\in D_i}(\theta_{LLM}^{(i-1)}), i=1,2,\ldots,n 
\end{equation}
where $\theta_{LLM}^{(0)}=\theta_{LLM}$. At the $i^{th}$ epoch, each data in $D_i$ forms a training instance with $\theta_{LLM}^{(i)}$ for model $R$, and is assigned with a difficulty score label of 0 since $\theta_{LLM}^{(i)}$ has mastered it. Correspondingly, data in $D_{i+1}$ is given a label of 1.

We intend to employ model $R$ to leverage both data and LLM's model feature. For input data $x\in[D_i, D_{i+1}]$, the initial model parameter $\theta_{LLM}$ is utilized to obtain data feature $z_1$. To acquire model features, we apply a trainable embedding matrix with Kaiming initialization, serving as an initial semantic segment. By processing the semantics through $\theta_{LLM}^{(i)}$, model feature $z_2$ is obtained. Then, two features are concatenated to form a hidden feature $z=\operatorname{concat}(z_1, z_2)$. This process is succinctly expressed as $H(x,\theta)=z$. Subsequently, we use another MLP structure to transform $z$ into a quantifiable difficulty score, that is, $R(x,\theta)=\operatorname{MLP}(z)=\operatorname{MLP}(H(x,\theta))$. The corresponding cross-entropy loss is calculated as:
\begin{equation}
\small
    \min_{R}L_R=E_{x\in D_i}[logR(x,\theta)]+E_{x\in D_{i+1}}[log(1-R(x,\theta))]
\label{eq:score}
\end{equation}
Additionally, we introduce an adversarial training mechanism to further enhance generalization of model $R$. In this process, discriminator $D$ engages in a rivalry with model $R$, progressively refining the capability of model $R$ in assessing data difficulty. Specifically, model $R$ tries to fool the discriminator to correctly predict whether the instance is easy. While the discriminator $D$ with a two-layer MLP structure is adversarially trained to accurately approximate the true difficulty label. Therefore, the training objectives for model $R$ and the discriminator $D$ are symmetrically formulated as:
\begin{equation}
\small
\begin{aligned}
\max_R L_R=E_{x\in D_i}[log R(x, \theta)] +E_{x\in D_{i+1}}[log R(x, \theta)]   
\end{aligned}
\end{equation}
\begin{equation}
\small
\begin{aligned}
\min_D L_D&=E_{x\in D_i}[log D(H(x, \theta)]\\
&+E_{x\in D_{i+1}}[log (1-D(H(x, \theta))]   
\end{aligned}
\end{equation}

\section{Experiments}
\subsection{Experiment Setting}
\noindent\textbf{Training Dataset} To intimate the educational scenario in which students simultaneously engage with multiple science curricula, we constructed a comprehensive instruction dataset for training, including mathematical reasoning, code generation, and general language understanding. They are sourced from 3 instruction datasets: training set of \textbf{GSM8K} \cite{cobbe2021training}, \textbf{Code Alpaca} \cite{chaudhary2023code}, and \textbf{ShareGPT}\footnote{\url{https://huggingface.co/datasets/anon8231489123/ShareGPT_Vicuna_unfiltered}} \cite{chiang2023vicuna}.

\noindent\textbf{Evaluation Benchmark and Metrics} We assess the aforementioned three capabilities using the following benchmarks and metrics: evaluation set of \textbf{GSM8K}, \textbf{HumanEval} \cite{chen2021evaluating}, and \textbf{MT-Bench} \cite{zheng2023judging}. The details about evaluation metrics are in Appendix \ref{sec:appendix_exp_set}.

\noindent\textbf{Baselines} We compare three categories of baseline methods, and all baselines used the same training dataset. The first category is efficient tuning methods based on \textbf{\textit{data selection}}, including \textbf{IFD} \cite{li2024quantity} and \textbf{DEITA} \cite{liu2024what}, in which DEITA represents the state-of-the-art methods.
The second category comprises efficient tuning methods based on \textbf{\textit{training order optimization}}. In addition to \textbf{random shuffle tuning} and \textbf{sequential tuning} on a dataset-by-dataset basis, \textbf{Tree-Instruct} \cite{zhao2024tree} and \textbf{Conifer} \cite{sun2024conifer}\footnote{We only used the metrics in the two papers to measure the data difficulty in our experiments.} are introduced as representatives of heuristic curriculum instruction tuning. \textbf{DMT} \cite{dong-etal-2024-abilities} is also a data order optimization baseline.

We also incorporated a third category of methods, which \textbf{\textit{individually train LLMs}} on three separate training datasets. This approach is designed to explore the impact of capability conflicts \cite{dong-etal-2024-abilities} embodied in various curricula (datasets). Additionally, we aim to assess whether CAMPUS can potentially mitigate these conflicts by adjusting curricula with the awareness of model competence changes, compared to other baselines.

\subsection{Main Experiments}
\begin{table*}[]
\centering
\renewcommand\arraystretch{1}
\setlength\tabcolsep{5pt}
\scalebox{0.65}{
\begin{tabular}{lccccccc}
\toprule
\textbf{Backbone LLM} & \multicolumn{3}{c}{\textbf{LLaMA-7B}} & \multicolumn{3}{c}{\textbf{LLaMA-13B}} & \textbf{Avg} \\ \cmidrule(lr){2-4} \cmidrule(lr){5-7}
Dataset               & GSM8K    & HumanEval    & MT-Bench    & GSM8K     & HumanEval    & MT-Bench    &              \\ \midrule
\multicolumn{8}{l}{\textit{Individual Training}}                                                                      \\
Math only             & \underline{35.56}    & -            & -           & \textbf{41.85}     & -            & -           & -        \\
Code only             &          & 11.59        & -           & -         & 12.20        & -           & -        \\
General only          & 9.55     & 7.93         & \underline{5.83}        & 12.28     & 9.15         & \underline{5.98}        & $26.17_{\pm0.13}$         \\ \midrule
\multicolumn{8}{l}{\textit{Data Selection}}                                                                           \\
IFD \cite{li2024quantity}                   &  33.43	& 13.40	& 5.69	& 37.62	& 15.24	& 5.78	& $35.73_{\pm0.07}$          \\
DEITA \cite{liu2024what}                & 33.21    & 14.02        & 5.73        & 37.83     & 15.24        & 5.88        & $36.07_{\pm0.15}$         \\ \midrule
\multicolumn{8}{l}{\textit{Training Order Optimization}}                                                              \\
Random Shuffle Tuning        & 35.03    & \underline{14.63}        & 5.63        & 40.11     & 15.24        & 5.76        & $36.49_{\pm0.11}$         \\
Sequential Tuning        & 32.45    & 13.41        & 5.65        & 36.62     & 15.85        & 5.71        & $35.32_{\pm0.14}$         \\
DMT \cite{dong-etal-2024-abilities}                  & 33.81    & 14.02        & 5.75        & 38.14     & 15.85        & 5.83        & $36.27_{\pm0.19}$        \\
Tree-Instruct \cite{zhao2024tree}         & 33.02	& 13.62	& 5.51	& 37.91	& 15.72	& 5.36	& $34.83_{\pm0.08}$            \\
Conifer \cite{sun2024conifer}            & 34.70     & 14.00           & 5.74        & 38.37                         & \underline{16.46} & 5.85 & $\underline{36.57}_{\pm0.11}$              \\
CAMPUS (ours)         & \textbf{35.86}    & \textbf{15.24}        & \textbf{5.95}        & \underline{40.56}     & \textbf{17.68}        & \textbf{6.01}        & $\textbf{38.16}_{\pm0.09}$        \\ \bottomrule
\end{tabular}}
\caption{The average results of three times for LLaMA-7B, 13B on three benchmarks. Since MT-Bench scores are in tenths, we multiplied them by 10 when calculating overall performance averages in the last column (the same below). See Appendix \ref{sec:llama33b} for performance on other models such as LLaMA-33B.}
\label{tab:main1}
\end{table*}

\begin{table*}[ht]
\centering
\renewcommand\arraystretch{1}
\setlength\tabcolsep{3.3pt}
\scalebox{0.65}{
\begin{tabular}{lcccccccccc}
\toprule
\textbf{Backbone LLM} & \multicolumn{3}{c}{\textbf{BLOOMZ-560M}} & \multicolumn{3}{c}{\textbf{BLOOMZ-1B7}} & \multicolumn{3}{c}{\textbf{BLOOMZ-3B}} & \multirow{2}{*}{\textbf{Avg}} \\ \cmidrule(lr){2-4} \cmidrule(lr){5-7} \cmidrule(lr){8-10}
Dataset               & GSM8K     & HumanEval     & MT-Bench     & GSM8K     & HumanEval     & MT-Bench    & GSM8K     & HumanEval    & MT-Bench    &                               \\ \midrule
\multicolumn{11}{l}{\textit{Individual Training}}                                                                                                                                   \\
Math only             & \textbf{7.81}      & -             & -            & \textbf{12.66}     & -             & -           & \textbf{15.24}     & -            & -           & -                         \\
Code only             & -         & 4.88          & -            & -         & 7.93          & -           & -         & 9.76         & -           & -                          \\
General only          & 0.68      & 3.05          & 3.26         & 1.36      & 3.66          & 3.78        & 2.33      & 6.10         & 4.11        & $14.30_{\pm0.17}$                           \\ \midrule
\multicolumn{11}{l}{\textit{Data Selection}}                                                                                                                                        \\
IFD  \cite{li2024quantity}                  &6.22	&4.27	&3.28	&9.37	&7.93	&3.6	&13.56	&10.97	&4.05	& $17.96_{\pm0.11}$                         \\
DEITA \cite{liu2024what}                & 6.60      & 4.88          & 3.24         & 10.24     & \underline{9.76}          & 3.71        & 13.65     & 11.59        & 4.13        & $18.61_{\pm0.09}$                           \\ \midrule
\multicolumn{11}{l}{\textit{Training Order Optimization}}                                                                                                                           \\
Random Shuffle Tuning        & \underline{7.43}      & 4.88          & 3.14         & \underline{11.37}     & 7.93          & 3.61        & 14.94     & 11.59        & 3.94        & $18.34_{\pm0.11}$                          \\
Sequential Tuning        & 6.37      & 4.88          & 3.15         & 9.55      & 8.54          & 3.65        & 13.42     & 11.59        & 4.09        & $18.14_{\pm0.11}$                           \\
DMT  \cite{dong-etal-2024-abilities}                    & 6.98      & \underline{5.49}          & \underline{3.28}         & 9.86      & 8.54          & \underline{3.81}        & 13.95     & \underline{12.80}        & \underline{4.25}        & $\underline{19.00}_{\pm0.09}$                           \\
Tree-Instruct \cite{zhao2024tree}        & 6.19	& 4.39	& 3.14	& 9.41	& 8.23	& 3.74	& 14.03	& 11.49	& 4.09	& $18.16_{\pm0.13}$                         \\
Conifer \cite{sun2024conifer}              & 6.79	& 4.88	& 3.28	& 9.89	& 8.54	& 3.8	& 13.83	& 11.59	& 4.16	& $18.66_{\pm0.10}$                            \\
CAMPUS (ours)         & 7.06      & \textbf{6.10}          & \textbf{3.43}         & 10.84     & \textbf{10.37}         & \textbf{4.13}        & \underline{15.09}     & \textbf{13.41}        & \textbf{4.46}        & $\textbf{20.34}_{\pm0.08}$      \\ \bottomrule                    
\end{tabular}}
\caption{The results of BLOOMZ-560M, 1B7, 3B on three benchmarks.}
\label{tab:main2}
\vspace{-0.5cm}
\end{table*}


We employ LLMs from the LLaMA \cite{touvronllama} and BLOOMZ \cite{muennighoff2023crosslingual} families as backbone models. The main results are shown in Table \ref{tab:main1} and \ref{tab:main2}. To validate the robustness of our results, we include the standard deviation of the average performance. We can observe that:

(1) Our CAMPUS significantly outperforms other baselines, including efficient instruction tuning methods based on data selection and training order optimization. Taking LLaMA as an instance, CAMPUS achieves an average gain of 7.0\% over other curriculum instruction tuning baselines Tree-Instruct and Conifer, which rely on a single heuristic difficulty. The improvement can be attributed to two aspects: on the one hand, CAMPUS dynamically selects the most suitable sub-curriculum during the training process based on LLM capability and data difficulty. On the other hand, its difficulty metrics are also model competence-aware, enabling flexible adjustments to the whole curriculum schedule.

(2) It is noted that existing curriculum instruction tuning did not even outperform random shuffle tuning on partial benchmarks, despite performing well under specific conditions reported in their original studies. This underperformance demonstrates their limited generalization, primarily due to their reliance on static metrics. For example, Tree-Instruct utilized the number of nodes in semantic tree formed by instruction as the difficulty metric, which may not be applicable to code instruction data, as its difficulty is not necessarily related to the code length.

(3) We find that some baselines trained on the full training set exhibit varying degrees of ``catastrophic forgetting'' compared to individual training, which is caused by the capability conflicting of the different training sets. Especially in mathematical reasoning, since both training and test sets are derived from the same GSM8K benchmark, the mixture of multiple datasets results in a significant degradation in mathematical capability. In contrast, CAMPUS shows marginal performance degradation, and in some cases, surpasses the performance of training solely on math data. This demonstrates that CAMPUS is capable of mitigating such conflict by perceiving the model capability changes on the fly and dynamically planning the most ``suitable'' sub-curriculum for the current LLMs.

(4) Efficient instruction tuning based on data selection are originally intended to enhance the LLM performance by filtering out harmful data. However, these methods also employ heuristic algorithms, which rendering them less generalizable. In the experiments, they also did not outperform random shuffle tuning. Instead of improving the performance of the model with reduced training data, it turns into a sacrifice of performance for time-efficiency.

(5) CAMPUS leads to a more pronounced performance rise on larger LLMs compared to smaller ones. This also emphasizes the importance of efficient instruction tuning for nowadays LLMs with increasingly larger parameter sizes.

\subsection{Ablation Experiments on CAMPUS}
\subsubsection{Ablation on Difficulty Metrics}

\begin{table}[]
\centering
\renewcommand\arraystretch{1}
\setlength\tabcolsep{3.3pt}
\scalebox{0.65}{
\begin{tabular}{lcccc}
\toprule
\textbf{Backbone LLM} & \multicolumn{3}{c}{\textbf{LLaMA-7B}} & \multirow{2}{*}{\textbf{Avg}} \\ \cmidrule{2-4}
Metrics               & GSM8K    & HumanEval    & MT-Bench    &                               \\ \midrule
\begin{tabular}[c]{@{}l@{}}Random Shuffle\\ Tuning\end{tabular}                 & 35.03    & 14.63        & 5.63        & 35.32                         \\ \midrule
Length $d_1$                & 33.36    & 14.63        & 5.74        & 35.12                         \\
MTLD $d_2$                    & 33.43    & 15.85        & 5.73        & 35.51                         \\
Loss $d_3$                    & 34.72    & 14.02        & 5.69        & 35.21                         \\ 
Scoring Model $d_4$                    & 35.48    & 15.85        & 5.91        & 36.82                         \\ \midrule
CAMPUS                & 35.86    & 15.24        & 5.95        & 36.87                        \\ \bottomrule
\end{tabular}}
\caption{Ablation experiments on difficulty metrics.}
\vspace{-0.4cm}
\label{tab:ablation_metrics}
\end{table}
Table \ref{tab:ablation_metrics} illustrates the results using a single difficulty metric, which is one of the difficulty metrics for the CAMPUS. The benefits from curriculum tuning based on single difficulties are relatively modest compared to the multiple difficulty-based CAMPUS. Static metrics, such as data length $d_1$, even have negative impacts on GSM8K and HumanEval, highlighting the limited generalization of conventional heuristic difficulties. This also confirms that data length is not a suitable metric for assessing the complexity of mathematical and coded data. Additionally, the scoring model we developed demonstrates promising performance, which is close to CAMPUS overall.
 
\subsubsection{Comparison on Different Schedulers}

\begin{table}[]
\centering
\renewcommand\arraystretch{1}
\setlength\tabcolsep{3.3pt}
\scalebox{0.65}{
\begin{tabular}{lcccc}
\toprule
\textbf{Backbone LLM} & \multicolumn{3}{c}{\textbf{LLaMA-7B}} & \multirow{2}{*}{\textbf{Avg}} \\ \cmidrule{2-4}
Scheduler             & GSM8K    & HumanEval    & MT-Bench    &                               \\ \midrule
Random                & 34.23    & 14.29        & 5.59        & 34.81                       \\
Sequential          & 34.26    & 14.37        & 5.54        & 34.68                          \\
PPL max               & 34.41    & 14.63        & 5.61        & 35.05                         \\ \midrule
PPL min               & 35.86    & 15.24        & 5.95        & 36.87                        \\ \bottomrule
\end{tabular}}
\caption{Comparison experiments on different sub-curriculum scheduling strategies.}
\label{tab:scheduler}
\vspace{-0.5cm}
\end{table}

We compare different sub-curriculum scheduling strategies in Table \ref{tab:scheduler}, including random selection, sequential selection from 1 to $n$, and both maximum and minimum PPL (used in CAMPUS). It is evident that selecting the sub-curriculum with the minimum PPL is the optimal scheduling strategy, which is in line with the curriculum learning principles and human intuition. Sub-curriculum with minimal perplexity implies that the LLM is primed to comprehend it, and thus the associated competencies of them can be more easily internalized by LLM. In contrast, sequential selection of sub-curriculum disrupts the ordered progression of the curriculum, as the difficulty of sub-curricula across different schedules cannot be directly compared. Consequently, the performance of sequential selection strategy is inferior even to random selection.

\subsubsection{Comparison on Data with Different Difficulty}
\begin{table}[]
\centering
\renewcommand\arraystretch{1}
\setlength\tabcolsep{3.3pt}
\scalebox{0.65}{
\begin{tabular}{lcccl}
\toprule
\textbf{Backbone LLM} & \multicolumn{3}{c}{\textbf{LLaMA-7B}} & \multicolumn{1}{c}{\multirow{2}{*}{\textbf{Avg}}} \\ \cline{2-4}
Dataset               & GSM8K    & HumanEval    & MT-Bench    & \multicolumn{1}{c}{}                              \\ \midrule
Full \& Random        & 35.03    & 14.63        & 5.63        & 35.32                                             \\ \midrule
1/3 Easy              & 32.30    & 13.41        & 5.89        & 34.86                                             \\
1/3 Medium            & 33.21    & 13.41        & 5.83        & 34.96                                             \\
1/3 Hard              & 34.34    & 14.02        & 5.76        & 35.33                                             \\ \midrule
Full \& CAMPUS        & 35.86    & 15.24        & 5.95        & 36.87                                             \\ \bottomrule
\end{tabular}}
\caption{Data difficulty level experiments.}
\label{tab:part_data}
\vspace{-0.5cm}
\end{table}

Table \ref{tab:part_data} illustrates the impact of data with different difficulty levels. Specifically, we evenly divided the instruction data programmed by CAMPUS into three portions, and categorized them as easy, medium, and hard, according to their precedence order. The results indicate that no single data segment provides more significant gains to LLMs than CAMPUS trained with the complete data, underscoring that each portion contributes to the overall performance. Similar experiments are conducted in Tree-Instruct \cite{zhao2024tree}, and its experiments show that the gain brought by the hard sub-dataset exceeds that of curriculum learning with the complete dataset. This contrary experimental conclusion is attributed to the sub-optimization of Tree-Instruct, which does not fully stimulate the potential of curriculum learning. Unlike Tree-Instruct, CAMPUS employs dynamic, multi-perspective difficulty metrics that customize the curriculum specifically to the competence and learning needs of the particular LLM. Of course, referring to the middle portion of Table \ref{tab:part_data}, we can similarly conclude that harder data is more critical for LLMs, which aligns with the view of previous work \cite{xu2023wizardlm,lu2023instag}.

\subsection{Generalization of CAMPUS}
\subsubsection{Compatibility with Other Methods}
\begin{table*}[h]
\centering
\setlength\tabcolsep{5pt}
\scalebox{0.55}{
\begin{tabular}{lccccccccc}
\toprule
\textbf{Backbone LLM} & \multicolumn{4}{c}{\textbf{LLaMA-7B}} & \multicolumn{4}{c}{\textbf{LLaMA-13B}} & \multirow{2}{*}{\textbf{Avg}} \\ \cmidrule(lr){2-5} \cmidrule(lr){6-9}
Dataset               & GSM8K    & HumanEval    & MT-Bench  &  Training Time (h)  & GSM8K     & HumanEval    & MT-Bench    &  Training Time (h)   &                         \\ \midrule
Random Shuffle Tuning  & 35.03  & 14.63  & 5.63  & 47  & 40.11  & 15.24  & 5.76   & 53  & 36.49 \\ \midrule
IFD                  &  33.43	& 13.40	& 5.69	& 37 + 8  & 37.62	& 15.24	& 5.78  & 37 + 10  	& 35.73                            \\
+   CAMPUS            & 34.77    & 14.63        & 5.79   & 17     & 38.49     & 15.85        & 5.85     & 20    & 36.69                        \\ \midrule
DEITA              & 33.21    & 14.02        & 5.73    & 41 + 4    & 37.83     & 15.24        & 5.88    & 41 + 5    & 36.07                             \\
+   CAMPUS            & 34.93    & 14.63        & 5.83    & 9    & 39.21     & 16.46        & 5.94    & 11    & 37.16                         \\ \midrule
CAMPUS                & 35.86    & 15.24        & 5.95    &  54   & 40.56     & 17.68        & 6.01    & 64    & 38.16             \\ \bottomrule               
\end{tabular}}
\caption{Experiments on our CAMPUS combined with other data selection-based efficient tuning methods. $x+y$ in training time for data selection methods means data selecting and LLM training time, respectively.}
\label{tab:plus_ours}
\vspace{-0.5cm}
\end{table*}

Our CAMPUS framework is designed to be compatible with various data selection-based efficient tuning methods, further enhancing the performance of LLMs. To verify the effectiveness of CAMPUS combined with them, we first filter the original training set leveraging the data selection-based method, and then apply our CAMPUS to the refined dataset. The results in Table \ref{tab:plus_ours} demonstrate that CAMPUS further boosts the performance of both IFD and DEITA, confirming its excellent adaptability. 
Moreover, while the selection-based method reduces the training time for LLMs, it simultaneously requires considerable time for selecting. Interestingly, the performance gain from CAMPUS is more pronounced with DEITA, which retains more data compared to IFD. This observation suggests that DEITA and IFD may inadvertently discard potentially useful data. Inspired by this, we believe that future work could combine the ideas of several types of efficient tuning methods in a unified way, and propose a more comprehensive data management strategy, considering more aspects such as data order, ratio, and the criteria for data exclusion, etc. This could potentially maximize the synergistic effects of combining different tuning methods and further enhance LLM training efficacy.

\subsection{Training Process of CAMPUS}
\subsubsection{Data Composition at Different Stages}
\begin{figure}[th]
\centering
\setlength{\abovecaptionskip}{0.1em}
\subfigure
{ \label{fig:data_percent1}
\includegraphics[width=0.8\columnwidth]{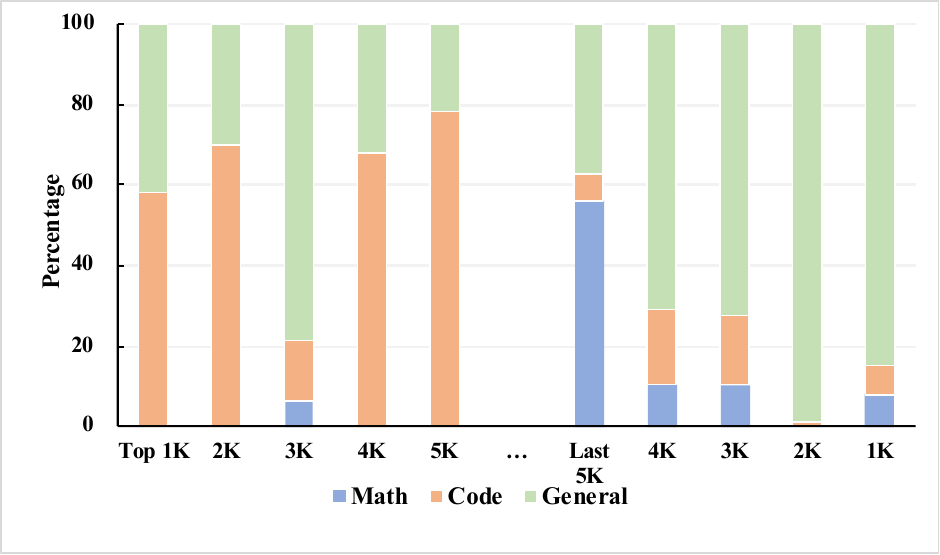} 
} 
\subfigure
{ \label{fig:data_percent2}
\includegraphics[width=0.8\columnwidth]{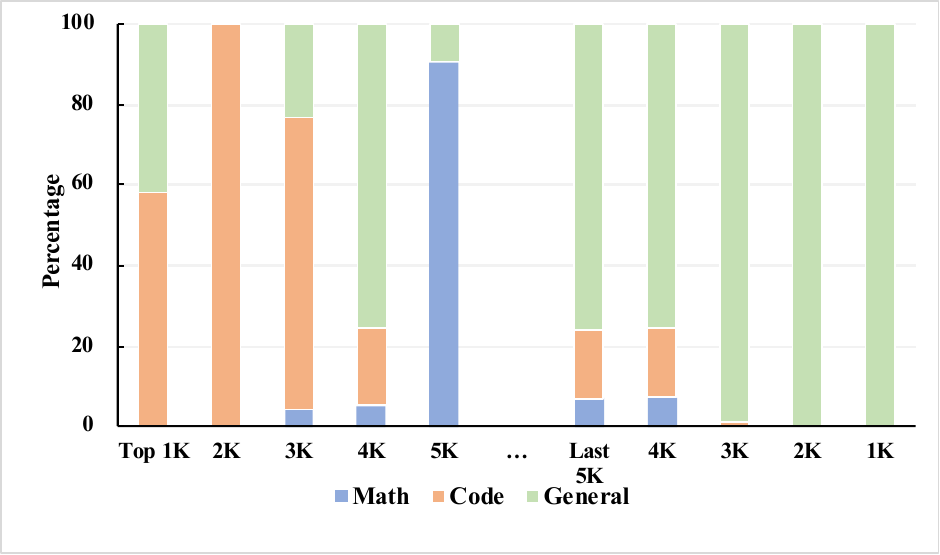} 
} 
\caption{Data composition of first/last 5K data by CAMPUS. Top: LLaMA-7B. Bottom: LLaMA-13B.} 
\label{fig:data_percent} 
\vspace{-0.5cm}
\end{figure} 

Figure \ref{fig:data_percent} reports the data type composition of the first and last 5K data scheduled by CAMPUS for different LLMs. Variations in data composition across different LLMs and training phases can be clearly observed. During the early stages of training, the code data proportion is relatively high, which may arise from the fact that LLMs require the underlying reasoning capabilities embodied in the code tasks, which is also indicated in some works \cite{bi2024program,chengbinding}. In contrast, the later stages of training witness an increase in math and general instruction data. Further case analysis in Table \ref{tab:case1} and \ref{tab:case2} in Appendix reveals that these later stages involve a significant number of mathematical tasks that require complex logical reasoning and intricate multi-round conversation data which proves that CAMPUS behaves in line with our intuition. As for the different LLMs, it is notable that math data are introduced earlier in the training process for larger LLMs (LLaMA-13B). Since larger LLMs possess more advanced foundational capabilities, allowing them to tackle complex mathematical tasks more readily and reducing their perceived difficulty more quickly.
This phenomenon again confirms the CAMPUS's ability to perceive the model's competence and tailor the LLM-specific curriculum dynamically.

\section{Conclusion}
In this paper, to address the rigidity issue of existing curriculum instruction tuning methods, we propose a \textbf{C}ompetence-\textbf{A}ware \textbf{M}ulti-\textbf{P}erspective c\textbf{U}rriculum in\textbf{S}truction tuning framework termed \textbf{CAMPUS}. CAMPUS offers the following advantages: (1) dynamic selection for sub-curriculum. (2) competency-aware adjustment to the curriculum schedule. (3) multiple difficulty-based curriculum schedule. Extensive experiments prove the superior performance of CAMPUS, compared to state-of-the-art efficient instruction tuning methods.

\clearpage
\section*{Limitations}
In this paper, we discuss efficient instruction tuning methods based on training order optimization, aiming to enable models to achieve higher performance using the same training data. The potential for combining with other types of efficient instruction tuning methods was primarily explored in our experiments. Specifically, we integrated our CAMPUS with the data selection based methods DEITA and IFD. The results demonstrate the feasibility of this idea, and future work could further explore how to generalize the ideas of different efficient instruction tuning methods in a unified way. For example, by drawing on the idea of reinforcement learning to view a data-wise management strategy as an action of agent.


\bibliography{custom}

\clearpage
\appendix
\section{Appendix for CAMPUS Method}
\label{sec:appendix}

\subsection{Supplementary Description of Difficulty Metrics for CAMPUS}
\label{sec:appendix_metrics}
In our experiments, we employed four difficulty metrics (i.e., $n$=4), of which loss and reward score are competence-aware. These metrics not only assess the inherent difficulty of the instruction data but also adjust according to the LLM's proficiency with the skills associated with the instruction data. In other words, they serve as difficulty indicator functions $d_i=f_i\left(\mathrm{inst},\theta_{\mathrm{LLM}}\right)$ with data and model parameter as inputs. Additionally, we introduced heuristic difficulty metrics data length and textual lexical diversity, considering their empirical effectiveness.

\noindent\textbf{Data Length} $d_1$ The data length is well-suited as a heuristic difficulty metric, aligning with intuitive human assessments: shorter instruction data are typically easier for LLMs to comprehend and learn. Conversely, longer instruction data often occurs in multi-round conversations, indicating that the instructions contain more diverse information and complex interactions requiring extended processing, thus making them more challenging to learn. Consequently, we employed the total length of the instruction and output tokens, denoted as $\operatorname{len}(x+y)$, as $d_1$. From the perspective of data length, it is natural for LLMs to first master basic skills from simple single-round conversations, and then learn to comprehensively utilize these skills in complex multi-round conversations.

\noindent\textbf{Measure of textual lexical diversity (MTLD)} $d_2$ Lexical diversity refers to the variety of different words employed in a text, which serves as a crucial factor in studies of linguistic complexity. MTLD \cite{mccarthy2010mtld} is a widely used quantitative measure for assessing lexical diversity. It is calculated as the mean length of sequential token lists in a text that maintains a given type-token ratio (TTR) threshold (0.72 in our paper). The TTR is defined as $\frac{l}{m}$, where $l$ and $m$ are the token length and the number of unique token types, respectively. A higher MTLD indicates greater vocabulary diversity in instruction data, which may contain more advanced vocabulary, or varied word combinations to convey the same meanings, which is more difficult to be learned by the LLMs. We selected MTLD as a metric due to its length-insensitivity compared to traditional lexical diversity metrics such as TTR. Text length does not always accurately reflect the data complexity. For instance, in programming code, lengthy text may include blocks with duplicated functionality, leading to an artificially inflated complexity solely based on length. Therefore, MTLD can be considered as a complementary perspective on instruction complexity that is distinct from data length.

\noindent\textbf{Data Loss} $d_3$ The data difficulty as perceived by LLMs depends not only on the data's inherent complexity, but also on the LLMs' comprehension abilities. Therefore, in addition to the two heuristics mentioned above, we introduce a simple, model competence-aware difficulty metric, data loss, as $d_3$. Data loss indirectly quantifies problem-solving capacity of LLM by measuring the difference between the predicted content and the ground-truth value, which is defined as:
$$
d_3=L(x,y,\theta_{LLm})=-\sum_{t=1}^N log p_{\theta_{LLM}}(y_t|x,y_{<t})
$$
where $N$ is the length of the output text sequence and $\theta_{LLm}$ is the LLM parameter. Predicting the next token $y_t$ requires utilizing the input $x$ and the preceding output sequence $y_{<t}$. A higher loss suggests that LLM struggles with tasks related to specific instruction data. The difficulty stems from either the intricate knowledge inherent in the data or the models' limited comprehension and reasoning capabilities. By monitoring the loss, we can orderly manage the instructional data.

\section{Appendix for Experiment Setting}
\label{sec:appendix_exp_set}
\subsection{Training Dataset}
\label{sec:traindata_appendix}
To imitate the educational scenario in which students simultaneously engage with multiple science curricula, we constructed a comprehensive instruction dataset for training, including mathematical reasoning, code generation, and general language understanding. They are sourced from the following instruction datasets, respectively:
\begin{itemize}[fullwidth,itemindent=1em]
    \item \textbf{GSM8K} \cite{cobbe2021training} is a mathematical dataset consisting of $\sim$8.5K high-quality grade school math problems that require multi-step reasoning. The dataset is divided into 6K training and 2K test instances.
    \item \textbf{Code Alpaca} \cite{chaudhary2023code} aims to build an instruction-following LLaMA model for code generation, which contains 20K instruction-following data generated by the techniques in the Self-Instruct.
    \item \textbf{ShareGPT}\footnote{\url{https://huggingface.co/datasets/anon8231489123/ShareGPT_Vicuna_unfiltered}} \cite{chiang2023vicuna} is a multi-round human-machine dialog dataset comprising 90K human queries and responses from ChatGPT or other chatbots. We cleaned it to reduce the size to 53K.
\end{itemize}

\subsection{Evaluation Benchmark and Metrics} 
We assess the aforementioned three capabilities of LLMs using the following benchmarks and metrics:
\begin{itemize}[fullwidth,itemindent=1em]
\item The test set of \textbf{GSM8K} is utilized to evaluate the mathematical reasoning ability of LLMs, which comprises 3K data. Following \citet{yuan2023scaling}, we use the accuracy of answers generated by greedy decoding (i.e., maj@1) as our metrics.
\item \textbf{HumanEval} \cite{chen2021evaluating} is the benchmark for code-writing with 164 handwritten programming problems, with an average of 7.7 tests per problem. Similar to \cite{chen2021evaluating}, we measure pass@$k$ on the HumanEval, where correctness is defined by an unbiased estimate of passing a set of unit tests within $k$ samples.
\item \textbf{MT-Bench} \cite{zheng2023judging} is a challenging benchmark widely adopted to assess the general instruction-following ability. It features multi-turn conversations across 8 domains. GPT-4 is employed as the judge to score model responses on a ten-point scale.
\end{itemize}

\subsection{Baselines} We compare three categories of baseline methods. The first category encompasses efficient tuning methods based on \textbf{\textit{data selection}}. These methods primarily employ metrics to assess the benefit of raw instruction data for LLMs from perspectives of quality, diversity, and complexity, and then discarding harmful data. \textbf{IFD} \cite{li2024quantity} and \textbf{DEITA} \cite{liu2024what} are included, where DEITA represents the state-of-the-art methods. 
The second category comprises efficient tuning methods based on \textbf{\textit{training order optimization}}, which enhance the efficiency by adjusting the training order of original instruction data. In addition to \textbf{random shuffle tuning} and \textbf{sequential tuning} on a dataset-by-dataset basis, \textbf{Tree-Instruct} \cite{zhao2024tree} and \textbf{Conifer} \cite{sun2024conifer}\footnote{We only used the metrics in the two papers to measure the data difficulty in our experiments.} are introduced as representatives of heuristic curriculum instruction tuning. \textbf{DMT} \cite{dong-etal-2024-abilities} is also a data order optimization baseline.

We also incorporated a third category of methods, which \textbf{\textit{individually train LLMs}} on three separate training datasets. This approach is designed to explore the impact of capability conflicts \cite{dong-etal-2024-abilities} embodied in various curricula (datasets). Additionally, we aim to assess whether CAMPUS can potentially mitigate these conflicts by adjusting curricula with the awareness of model competence changes, compared to other baselines.

\subsection{Training Details}
The CAMPUS is implemented by LLaMA-Factory framework\footnote{\url{https://github.com/hiyouga/LLaMA-Factory}} and fine-tuned with full parameters. The temperature is configured to 0.95, and the top-k is set to 50, indicating that each generation step involves sampling from 50 candidate tokens. The learning rate is managed with a 0.03 warm-up ratio, ramping up linearly to 2e-5 during the initial 3\% of the training data. Furthermore, we implement a cosine annealing strategy to adjust the learning rate in a cyclical manner, helping the model converge on the optimal solution.

When training the reward model $R$ and the discriminator $D$, we choose a batch size of 4 and a learning rate of 1e-5. Both models utilize identical MLP structures to map the hidden feature $z$ into a score, with a hidden dimension of 256. Employing the Kaiming initialization technique, they undergo continuous training through stochastic gradient descent (SGD), operating iteratively without parameter sharing. To accelerate convergence, we iterate twice for the training of $R$ and $D$ after each fine-tuning of the LLM.

\section{Supplementary Experiments on Generalization of CAMPUS}
\subsection{Performance Differences on Different Domains of MT-Bench}
\begin{table*}[]
\centering
\renewcommand\arraystretch{1}
\setlength\tabcolsep{3.3pt}
\scalebox{0.9}{
\begin{tabular}{lcccccc}
\toprule
Backbone LLM$\rightarrow$                                             & \multicolumn{3}{c}{\textbf{LLaMA-7B}} & \multicolumn{3}{c}{\textbf{LLaMA-13B}} \\ \cmidrule(lr){2-4} \cmidrule(lr){5-7} 
\begin{tabular}[c]{@{}l@{}}Reward Model\\ Trained on $\downarrow$ \end{tabular} & GSM8K    & HumanEval    & MT-Bench    & GSM8K     & HumanEval    & MT-Bench    \\ \midrule
\textbf{LLaMA-7B}                                                          & 35.48    & 15.85        & 5.91        & 37.86     & 16.46        & 5.96        \\
\textbf{LLaMA-13B}                                                         & 38.61    & 16.89        & 5.95        & 40.56     & 17.68        & 6.01       \\ \bottomrule
\end{tabular}}
\caption{Experiments on the generalization of scoring models.}
\label{tab:reward_gener}
\end{table*}

\begin{figure}[t]
\centering
\scalebox{0.9}{
\includegraphics[width=0.48\textwidth]{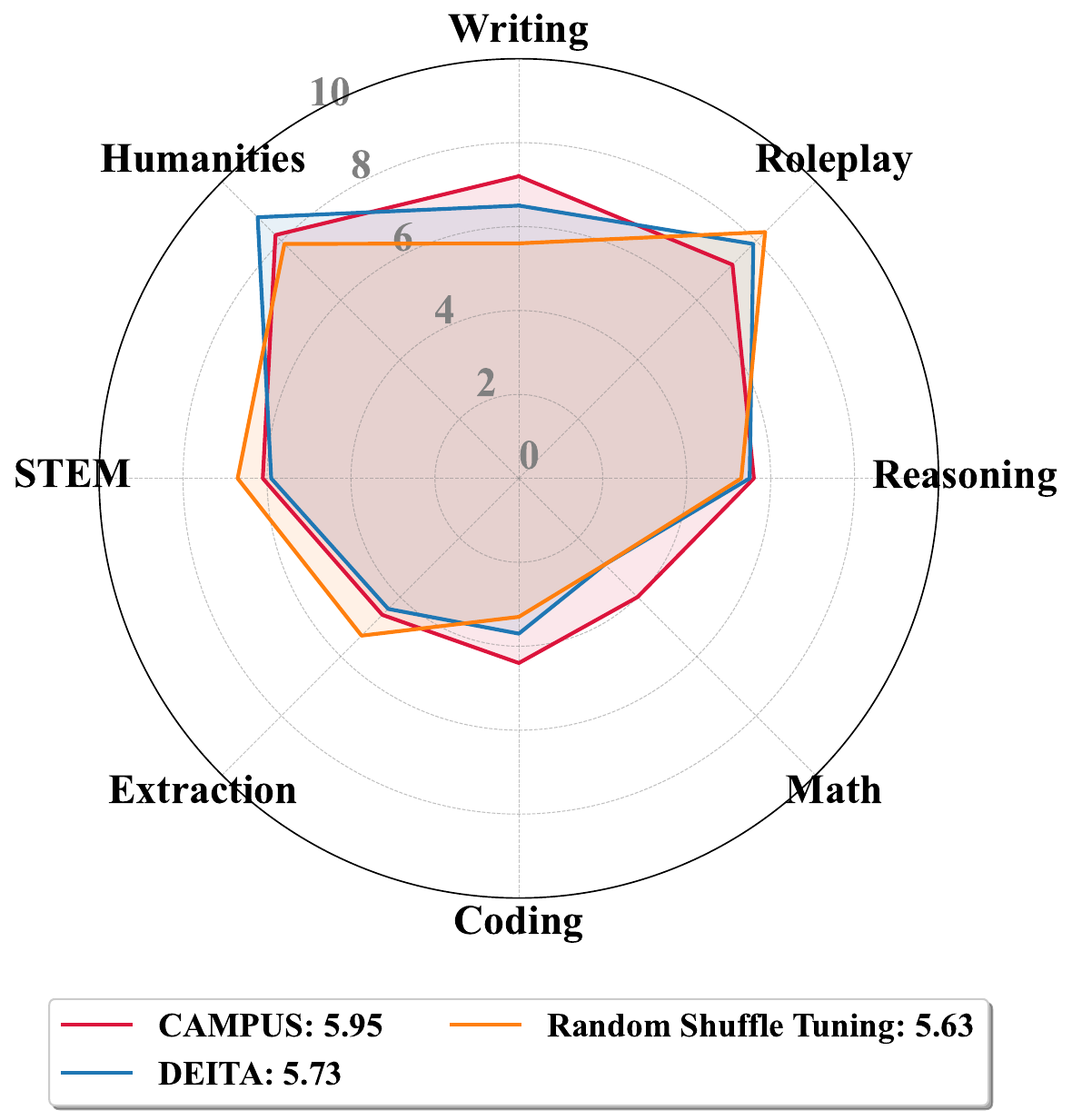}}
\caption{Model performance comparison on different domains of MT-Bench.}
\label{fig:radar}
\end{figure}

\begin{table*}[t]
\centering
\renewcommand\arraystretch{1}
\setlength\tabcolsep{3.3pt}
\scalebox{0.75}{
\begin{tabular}{lccccccc}
\toprule
\textbf{Backbone LLM} & \multicolumn{3}{c}{\textbf{LLaMA-7B}} & \multicolumn{3}{c}{\textbf{LLaMA-13B}} & \multirow{2}{*}{\textbf{Avg}} \\ \cmidrule(lr){2-4} \cmidrule(lr){5-7} 
\textbf{Method}               & GSM8K     & HumanEval     & MT-Bench     & GSM8K     & HumanEval     & MT-Bench    &   \\ \midrule
CAMPUS & 35.86 & 15.24 & 5.95 & 40.56 & 17.68 & 6.01 &  \textbf{38.16} \\
+ Label Smoothing \& Upsampling & 36.39 & 15.93 & 5.99 & 41.22 & 18.03 & 6.09  & \textbf{38.73} \\ \bottomrule               
\end{tabular}}
\caption{The performance of CAMPUS with label smoothing and upsampling on LLaMA-7B and LLaMA-13B.}
\label{tab:score_method}
\end{table*}

In Figure \ref{fig:radar}, we compare the performance of different models across the eight domains of MT-Bench, as represented in a radar chart. The chart demonstrates that CAMPUS covers a relatively larger area, indicating a more comprehensive and stable performance across all domains. While DEITA and random shuffle tuning exhibit superior performance in specific domains, CAMPUS consistently maintains robust performance without significant drawbacks in any particular area. In contrast, the other two baseline methods display pronounced deficiencies in challenging domains such as math, writing, and code. 
We attribute the stability of CAMPUS to its ability to perceive whether the abilities corresponding to different training data can be internalized by the current LLM. This capability allows CAMPUS to select the most appropriate training data tailored to the needs of the LLM, rather than employing a blind selection. This strategic selection process ensures that CAMPUS enhances LLM capabilities uniformly across diverse domains, preventing significant performance drops in any single area and promoting a well-rounded skill set in the model.

\subsection{Generalization of Scoring Model}
In this section, we explore the generalization capabilities of the scoring model R when trained with different LLMs. Table \ref{tab:reward_gener} assesses the curriculum scheduling ability of scoring models obtained by training with LLaMA-7B and 13B, respectively. The results, particularly evident from the first row, reveal that the scoring model trained with the smaller model (LLaMA-7B) can also effectively aid in the training of the larger model (LLaMA-13B). This transfer ability is highly desired in industrial scenarios. When computational resources are tight, instead of developing a new scoring model from scratch to perfectly align with the target LLM, organizations can employ an existing scoring model as an off-the-shelf dynamic curriculum schedule model directly for optimizing the training strategy of the target LLM.

Besides, we introduce two improvements to the scoring model to enhance its effectiveness: (1) applying label smoothing to the loss in Equation \ref{eq:score} to overcome overfitting, and (2) performing upsampling to mitigate label imbalance. Table \ref{tab:score_method} shows that these refinements to scoring model lead to further performance gains.

\subsection{Generalization on other benchmarks}
To further demonstrate the generalization of CAMPUS, we extend our evaluation to three additional widely used benchmarks: MATH~\cite{hendrycks2021measuring} for mathematics reasoning, MBPP~\cite{austin2021program} for code generation, and MMLU~\cite{hendrycksmeasuring} for general instruction following. As shown in Table \ref{tab:other_benchmarks}, CAMPUS consistently outperforms other state-of-the-art methods on these benchmarks, illustrating strong generalization ability.

\begin{table}[ht]
\centering
\renewcommand\arraystretch{1}
\setlength\tabcolsep{9.3pt}
\scalebox{0.75}{
\begin{tabular}{lccccccc}
\toprule
\textbf{Method}               & MATH     & MBPP     & MMLU   \\ \midrule
DEITA & 14.71 & 20.29 & 40.52 \\
Random Shuffle Tuning & 15.33 & 20.45 & 38.23  \\
DMT & 15.52 & 20.87 & 42.41  \\
Conifer & 15.28 & 20.62 & 41.57  \\ \midrule
\textbf{CAMPUS(ours)} & \textbf{16.24} & \textbf{21.71} & \textbf{44.18}  \\ \bottomrule               
\end{tabular}}
\caption{The performance of CAMPUS on three benchmarks with LLaMA-7B.}
\label{tab:other_benchmarks}
\end{table}

\section{Supplementary Experiments on Training Process of CAMPUS}
\subsection{Model Convergence During Training}
\begin{figure}[t]
\centering
\setlength{\abovecaptionskip}{0.5em}
\subfigure[Average performance curve of LLaMA-7B model during training process.] 
{ \label{fig:process1}
\includegraphics[width=0.95\columnwidth]{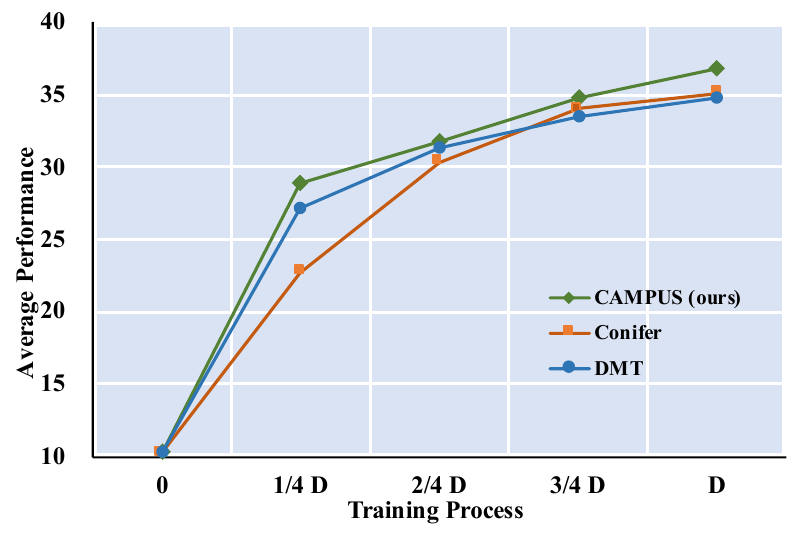} 
} 
\subfigure[Loss curve of LLaMA-7B model during training process.] 
{ \label{fig:process2}
\includegraphics[width=0.95\columnwidth]{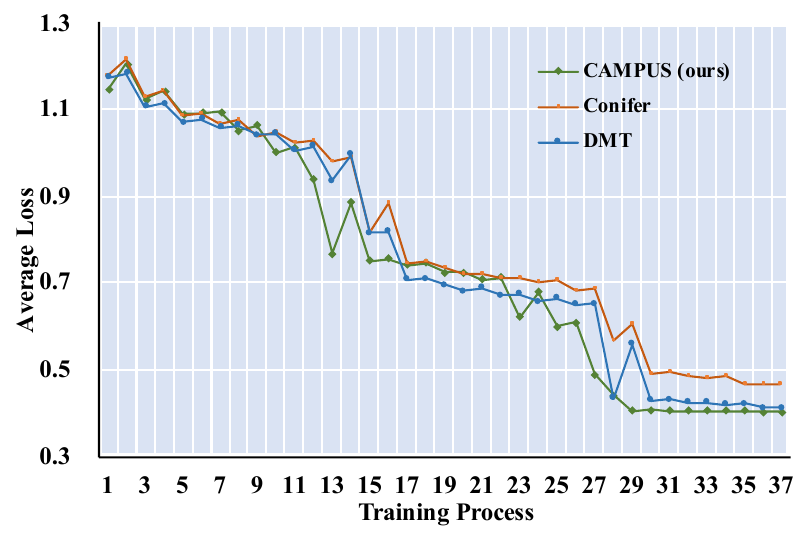} 
} 
\caption{Visualization of performance and loss.} 
\label{fig:process} 
\end{figure}

We visualize the training process for CAMPUS and the other two baselines in Figure \ref{fig:process}. Figure \ref{fig:process1} illustrates the average performance of the model after training with the corresponding amount of data, where D represents the full dataset. Figure \ref{fig:process2} records the average loss values at the corresponding step. We can observe that CAMPUS achieves faster loss convergence compared to the baseline model. Additionally, throughout the training process, CAMPUS consistently outperforms the baseline models and reaches higher performance upper bounds. This performance trend effectively demonstrates that CAMPUS aligns with the ultimate goal of curriculum instruction tuning, optimizing overall model performance, meanwhile facilitating learning efficiency.

\subsection{Supplementary Ablation on Difficulty Metrics}
\label{sec:append_metric}
The ablation experiments on difficult metrics in Table \ref{tab:append_ablation_metric} are complementary to those in Table \ref{tab:ablation_metrics}. In Table \ref{tab:append_ablation_metric}, we removed the individual metrics of CAMPUS to provide a different perspective on understanding the different metrics.
The trends observed are consistent with Table \ref{tab:ablation_metrics}. For example, removing the length metric improved GSM8K performance slightly, but overall, all metrics contributed to the performance gains of CAMPUS. These results support the complementary nature of the difficulty metrics.

\begin{table}[]
\centering
\scalebox{0.66}{
\begin{tabular}{lcccc}
\toprule
\textbf{Method}                & \textbf{GSM8K} & \textbf{HumanEval} & \textbf{MT-Bench} & \textbf{Avg}   \\ \midrule
Random Shuffle Tuning & 35.03 & 14.63     & 5.63     & 35.32 \\
CAMPUS                & 35.86 & 15.24     & 5.95     & 36.87 \\ \midrule
w/o Length $d_1$           & 36.39 & 14.98     & 5.83     & 36.56 \\
w/o MTLD $d_2$             & 36.41 & 14.93     & 5.68     & 36.05 \\
w/o Loss $d_3$              & 36.02 & 15.31     & 5.76     & 36.31 \\
w/o Scoring Model $d_4$     & 34.81 & 14.69     & 5.74     & 35.63 \\ \bottomrule
\end{tabular}}
\caption{Supplementary ablation experiments on difficulty metrics.}
\label{tab:append_ablation_metric}
\end{table}

\begin{table}[]
\centering
\small
\scalebox{0.73}{
\begin{tabular}{lcc}
\toprule
& \textbf{Training Time (Scoring Model)} & \textbf{Training Time (LLMs)} \\ \midrule
BLOOMZ 3B & 15 h (75\%) & 20 h \\ 
LLaMA 7B & 41 h (76\%) & 54 h \\
\bottomrule
\end{tabular}}
\caption{Training time comparison for both the scoring model and the LLMs.}
\label{tab:training_time}
\end{table}

\begin{table}[]
\centering
\scalebox{0.66}{
\begin{tabular}{lcccc}
\toprule
\textbf{Method}        & \textbf{GSM8K} & \textbf{HumanEval} & \textbf{MT-Bench}    &   \textbf{Avg}    \\ \midrule
DEITA & 47.43 & 21.44 & 6.22 & 43.69 \\
Random Shuffle Tuning & 46.23 & 17.32 & 5.93 & 40.95 \\
DMT & 49.51 & 22.62 & 6.45 & 45.54 \\
Conifer & 48.74 & 23.15 & 6.37 & 45.20 \\ \midrule
\textbf{CAMPUS (ours)} & \textbf{51.38} & \textbf{25.84}     & \textbf{6.79}    &   \textbf{48.37}    \\ \bottomrule
\end{tabular}}
\caption{The performance of CAMPUS and other baselines on LLaMA-33B.}
\label{tab:performance_llama33b}
\end{table}


\subsection{Training Time Analysis}
\label{sec:time}
We include a detailed analysis of training times for both the scoring model and the LLMs, as shown in Table \ref{tab:training_time}:

All baselines used the same training data. Hence, training time for training order optimization-based methods remains consistent. Data selection-based methods reduce time but sacrifice performance due to data over-deletion.

\subsection{CAMPUS Performance on Other LLMs}
\label{sec:llama33b}
To verify the performance of CAMPUS on larger or more recent LLMs, we conduct additional experiments on LLaMA-33B in Table \ref{tab:performance_llama33b}. As expected, CAMPUS demonstrated stable and superior performance compared to baselines, showcasing excellent scalability.

\begin{table*}[]
\centering
\scalebox{0.75}{
\begin{tabular}{llp{0.8\textwidth}}
\toprule
\textbf{Data Stage}                                                                     & \textbf{Data Type} & \textbf{Instruction Data} \\ \midrule
\multirow{3}{*}{\begin{tabular}[c]{@{}l@{}}First 1K \\Data\end{tabular}} & Math             &    \begin{tabular}[c]{@{}p{0.8\textwidth}@{}}
\#\#\# Instruction:\\Output the 5th Fibonacci number.\\\\\#\#\# Response:\\5. \end{tabular}                  \\ \cmidrule{2-3}
                                                                                   & Code             &  \begin{tabular}[c]{@{}l@{}}
\#\#\# Instruction:\\Create an HTML page with a text box that stores entered data in the local storage.\\\\
\#\#\# Response:\\<html>\\<head>\\  <title>Text Box</title>\\</head>\\\\<body>\\  <input type="text" id="text\_box" />\\</body>\\\\<script>\\  // Store data in localStorage\\  function storeData() \{\\\quad let inputValue = document.getElementById("text\_box").value;\\\quad localStorage.setItem("inputValue", inputValue);\\  \}\\\\  // Set the text box value from localStorage\\  window.onload = () => \{\\\quad if(localStorage.getItem("inputValue"))\\\quad   document.getElementById("text\_box").value = localStorage.getItem("inputValue");\\  \}\\\\  // Set up the event listener\\  document.getElementById('text\_box').addEventListener('input', storeData);\\</script>\\\\</html>          
                                                                                   \end{tabular}  \\ \cmidrule{2-3}
 & General             &    \begin{tabular}[c]{@{}p{0.8\textwidth}@{}}
\#\#\# Instruction:\\Hello.\\\\ \#\#\# Response:\\Hello! How can I help you today? \end{tabular}                      \\ \bottomrule
\end{tabular}}
\caption{The data cases sampled from the first 1K data.}
\label{tab:case1}
\end{table*}

\begin{table*}[]
\centering
\scalebox{0.75}{
\begin{tabular}{llp{0.8\textwidth}}
\toprule
\textbf{Data Stage}                                                                     & \textbf{Data Type} & \textbf{Instruction Data} \\ \midrule
\multirow{3}{*}{\begin{tabular}[c]{@{}l@{}}Last 1K \\Data\end{tabular}} & Math             &    \begin{tabular}[c]{@{}p{0.8\textwidth}@{}}
\#\#\# Instruction:\\
John buys a chair. He then buys a table that is 3 times the price of the chair. Then, he buys a couch that is 5 times the price of the table. If John paid \$380 for all these items, what is the price of the couch?\\
\\
\#\#\# Response:\\
Let x be the price of the chair.\\
The table costs 3*x.\\
The couch costs 3*x*5=15*x.\\
x+3*x+15*x=380\\
19*x=380\\
x=<<20=20>>20\\
The couch costs 20*15=<<20*15=300>>300 dollars. 
\end{tabular} \\ \cmidrule{2-3}
                                                                                   & Code             &  \begin{tabular}[c]{@{}p{0.8\textwidth}@{}}
\#\#\# Instruction:\\Generate a function in python that takes two lists and returns the product of their longest common substrings.\\\\
\#\#\# Input:\\ list1 = [1, 2, 3, 4, 5]\\list2 = [2, 3, 5, 7]\\\\
\#\#\# Response:\\ def longest\_common\_substring(list1, list2):\\\quad longest\_substring = 0\\\quad for i in range(len(list1)):\\\quad \quad for j in range(len(list2)):\\\quad \quad \quad if list1[i] == list2[j]:\\\quad \quad \quad \quad current\_substring = 1\\\quad \quad \quad \quad k, l = i+1, j+1\\\quad \quad \quad \quad while k < len(list1) and l < len(list2) and list1[k] == list2[l]:\\\quad \quad \quad \quad \quad current\_substring += 1\\\quad \quad \quad \quad \quad k += 1\\\quad \quad \quad \quad \quad l += 1\\\quad \quad \quad \quad longest\_substring = max(longest\_substring, current\_substring)\\\\\quad if longest\_substring == 0:\\\quad \quad return 0\\\quad else:\\\quad \quad return list1[0] * list2[0] * longest\_substring
                                                                                   \end{tabular}  \\ \cmidrule{2-3}
 & General             &    \begin{tabular}[c]{@{}p{0.8\textwidth}@{}}
\#\#\# Instruction:\\
5.0 Varying Magical Theories\\
5.1 Miseishun Theory; Spiritual Belief Method\\
The Miseishun hubworld has a comparatively spiritual take on arcanum, known as the Spiritual Belief Method (SBM), opposed to the theoretical views practiced in TOR, Theoretical Magical Formulae (TMF)...\\
\\
\#\#\# Response:\\
It seems that the "Looped Filter Theory" is a theory about...\\
\\
\#\#\# Instruction:\\
"6.0 Duinn\\
This section was written in coalition with TOR Combined Holy Clergy (TORCHC).\\
Despite the focus of this document being Arcanum, a necessary distinction to make in this report is that Duinn is not classified as magic...\\
\\
\#\#\# Response:\\
It seems that in this fictional universe, there are multiple types of magic and magical abilities that people can possess and use...\\
\\
.......\\
\\
More than 7 round conversations.
\end{tabular}                 \\ \bottomrule
\end{tabular}}
\caption{The data cases that sampled from the last 1K data.}
\label{tab:case2}
\end{table*}

\end{document}